\documentclass[runningheads]{llncs}
\usepackage{times}
\usepackage{latexsym}
\usepackage{amsmath}
\usepackage{amssymb}
\usepackage{graphicx}
\usepackage{booktabs}
\usepackage{multirow}
\usepackage{algorithm}
\usepackage{hyperref}
\usepackage{algorithmicx}  
\usepackage{algpseudocode}  
\usepackage{bbm}

\makeatletter
\newcommand{\printfnsymbol}[1]{\textsuperscript{\@fnsymbol{#1}}}
\def\@fnsymbol#1{\ensuremath{\ifcase#1\or * \or \dagger \else\@ctrerr\fi}}
\makeatother

\begin{document}
\newcommand\BibTeX{B\textsc{ib}\TeX}
\title{An Investigation on Different Underlying Quantization Schemes for Pre-trained Language Models} 
\titlerunning{An Investigation on Quantization Schemes}
%
\author{Zihan Zhao\inst{1,2}\thanks{Zihan Zhao and Yuncong Liu are co-first authors and contribute equally to this work.}
\and Yuncong Liu\inst{1,2}\printfnsymbol{1}
\and Lu Chen\inst{1,2}\thanks{The corresponding authors are Kai Yu and Lu Chen.}
\and Qi Liu\inst{1,2}
\and Rao Ma\inst{1,2}
\and Kai Yu\inst{1,2}\printfnsymbol{2}
}
\authorrunning{Z. Zhao et al.}
%
\institute{MoE Key Lab of Artificial Intelligence, AI Institute, Shanghai Jiao Tong University \and
SpeechLab, Department of Computer Science and Engineering Shanghai Jiao Tong University, Shanghai, China
\\
\email{\{zhao\_mengxin,assassain\_lyc\}@sjtu.edu.cn}
\\
\email{\{chenlusz,liuq901,rm1031,kai.yu\}@sjtu.edu.cn}}
\maketitle              
\begin{abstract}
Recently, pre-trained language models like BERT have shown promising performance on multiple natural language processing tasks. However, the application of these models has been limited due to their huge size. To reduce its size, a popular and efficient way is quantization. Nevertheless, most of the works focusing on BERT quantization adapted primary linear clustering as the quantization scheme, and few works try to upgrade it. That limits the performance of quantization significantly. In this paper, we implement k-means quantization and compare its performance on the fix-precision quantization of BERT with linear quantization. Through the comparison, we verify that the effect of the underlying quantization scheme upgrading is underestimated and there is a huge development potential of k-means quantization. Besides, we also compare the two quantization schemes on ALBERT models to explore the robustness differences between different pre-trained models.
\end{abstract}

\section{Introduction}
\indent Pre-trained transformer-based models \cite{transformer} recently have achieved state-of-the-art performance at a variety of natural language processing (NLP) tasks, such as sequence tagging and sentence classification. Among them, BERT models \cite{bert} based on transformer architecture \cite{transformer} have drawn even more attention because of their great performance and generality. However, the memory and computing consumption of these models are prohibitive. Even the relatively small versions of BERT models (e.g., BERT-base) contain more than 100 million parameters. The over-parameterized 
characteristic makes it challenging to deploy BERT models on devices with constrained resources, such as smartphones and robots. Therefore, compressing these models is an important demand in the industry. 
\\
\indent One popular and efficient method for model compression is quantization. To reduce model sizes, quantization represents the parameters of the model by fewer bits instead of the original 32 bits. 
With proper hardware, quantization could significantly reduce the memory footprint while accelerating inference. There have been many works focusing on quantizing models in the computer vision area \cite{binary:2,ternary:1,w1a2,han2015deep,hawq,haq}, while much fewer works have been done on NLP \cite{kaiyushi,raoma,efficient,zip,TQ}. Pilot works of transformer quantization include \cite{efficient,zip,TQ}. They successfully quantized transformer models to 8 or 4 bits while maintaining comparable performance. Moreover, to the best of our knowledge, there are only two published works focusing on BERT quantization \cite{q8bert,q-bert}. \cite{q8bert} applied 8-bit fixed-precision linear quantization to BERT models and achieved a compression ratio of 4$\times$ with little accuracy drop. \cite{q-bert} 
improved the quantization performances by group-wise mix-precision linear quantization based on the Hessian matrix of the parameter tensors. \\
\indent However, for the underlying quantization scheme, most of the above transformer quantization works, especially the BERT quantization works utilized linear clustering,
which is a primary clustering method. Although it can process fast and easily, the quantized results cannot represent the original data distribution well. As a result, \cite{q8bert} only manages to quantize BERT to 8 bits. Although the other BERT quantization work \cite{q-bert} has achieved much higher compress ratios without quantization scheme upgrading, the group-wise method they developed is rather time-consuming and increases the latency significantly. Although it is believed that replacing linear clustering with a better clustering method can improve the performance of quantized models.
The effect of the quantization scheme upgrading is rather underestimated.
Therefore, in this paper, we explore the effect of simply upgrading the quantization scheme from linear clustering to k-means clustering, and compare the performance of the two schemes. Furthermore, to see the effect on other pre-trained language models, we also compare the two quantization schemes on ALBERT models \cite{albert}, which is an improved version of BERT.\\
\indent In summary, we applied k-means and linear quantization on BERT and ALBERT and test their performances on GLUE benchmarks. Through this, we verify that simple upgrading of quantization scheme could result in great performance increases and simple k-means clustering has great potential as BERT quantization scheme. Moreover, we also show that the number of k-means iterations plays an important role in the k-means quantization. Through further comparison, we discover that ALBERT is less robust than BERT in terms of quantization, as the parameter sharing has reduced the redundancy of the parameters.




\section{Background: BERT and ALBERT}
\indent In this section, we briefly introduce the architectures of BERT and ALBERT models and point out the version of the models we used in our experiments. 

\subsection{BERT}\label{bert_arch}
\indent BERT models \cite{bert} are a special kind of pre-trained transformer-based network.
They mainly consist of embedding layers, encoder blocks, and output layers. 
There is no decoder block in BERT models. Each encoder block contains one self-attention layer (includes three parallel linear layers corresponding to query, key, and value) and 3 feed-forward layers (each includes one linear layer).\\
\indent For each self-attention layer, BERT utilize the multi-head technique to further improve its performance.
For the each self-attention head, there are 3 weight matrices $\mathbf{W}_q, \mathbf{W}_k,$ and $\mathbf{W}_v,$, where $\mathbf{W}_q, \mathbf{W}_k, \mathbf{W}_v\in \mathbb{R}^{d\times \frac{d}{h}}$ ($h$ is the number of heads in each self-attention layer). Let $X\in \mathbb{R}^{n\times d}$ denote the input of the corresponding self-attention layer. Therefore, the output of the self-attention head is calculated as:
\begin{equation}
    \begin{split}
    \mathbf{Q}=\mathbf{X}\mathbf{W}_{q} \indent \mathbf{K}=\mathbf{X}&\mathbf{W}_{k} \indent \mathbf{V}=\mathbf{X}\mathbf{W}_{v}\\
    \mathrm{Attention}(\mathbf{Q},\mathbf{K},\mathbf{V})=&\mathrm{softmax}(\dfrac{\mathbf{Q}\mathbf{K}^{T}}{\sqrt{d}})\mathbf{V}, 
    \end{split}
\end{equation}\par
Then, for each self-attention layer, the outputs of all its self-attention heads are concatenated sequentially to generate the output of the corresponding layer.\\
\indent Specifically, in our work, we use the bert-base-uncased version of BERT models, which has 12 encoder blocks and 12 heads for each self-attention layer, to carry out the following experiments.

\subsection{ALBERT}\label{albert_arch}
\indent Compared to BERT, ALBERT contributes three main improvements. First, ALBERT models decompose the embedding parameters into the product of two smaller matrices.
Second, they adapt cross-layer parameter sharing to improve parameter efficiency. These two improvements can significantly reduce the total number of parameters and make the model more efficient. Moreover, parameter sharing can also stabilize network parameters.
Third, they replace next-sentence prediction (NSP) loss with sentence-order prediction (SOP) loss while pre-training. This makes the models focus on modeling inter-sentence
coherence instead of topic prediction and improves the performance on multi-sentence encoding tasks.\\
\indent Specifically, in this paper, we use the albert-base-v2 version of ALBERT models, which also has 12 encoder blocks (where all parameters are shared across layers) and 12 heads for each self-attention layer. 

\section{Methodology}
\indent In this section, we first introduce the quantization process in our experiments (Section \ref{process}), then explain the two quantization schemes we used in detail (Section \ref{linear}, \ref{k-means}).
\subsection{Overview}\label{process}
\indent To compare linear and k-means quantization schemes on pre-trained transformer-based models, we test the performance of quantized models on different downstream tasks. Specifically, for each chosen task, the following experiments are carried out sequentially: fine-tuning the pre-trained models (BERT and ALBERT) on the downstream task; quantizing the task-specific model; fine-tuning the quantized model. Then the performance of the resulting model is tested on the validation set of each chosen task.\\
\indent To avoid the effect of other tricks, we simply apply the two quantization scheme (linear and k-means) following fix-precision quantization strategy without any tricks. We quantize all the weight of the embedding layers and the fully connected layers (except the classification layer). For each weight vector, after quantization, it will be represented by a corresponding cluster index vector and a centroid value vector, and each parameter of the weight vectors will be replaced with the centroid of the cluster which it belongs to.\\
\indent After the model is quantized, we further fine-tune it on the corresponding downstream tasks while maintaining quantized. For the forward pass, we reconstruct each quantized layer by its cluster index vector and centroid value vector. For the backward pass, while updating the rest parameters normally, we update the quantized parameters by training the centroids vectors. More specifically, the gradient of each parameter in the centroid vectors is calculated as the average of the gradients of the parameters that belong to the corresponding cluster. Then, the centroids value vectors are updated by the same back-propagation methods.

\subsection{Linear Quantization}\label{linear}
\indent Suppose that we need to quantize a vector $v$ to $k$ bits ($k$-bit quantization). We first search for its minimum value $v_{min}$ and maximum value $v_{max}$. The range $[v_{min}, v_{max}]$ is then divided into $2^{k}$ clusters with width
\begin{equation}
width=\frac{v_{max}-v_{min}}{2^k}.
\end{equation}
Define function $\hat{Q}$ as 
\begin{equation}
\hat{Q}(v_i)=\lfloor \frac{v_i-v_{min}}{width} \rfloor,
\end{equation}
whose value is between $0$ and $2^k-1$. Such that each parameter $v_i$ belongs to the $\hat{Q}(v_i)$-th cluster. And $v_i$ will be replaced with the centroid of $\hat{Q}(v_i)$-th cluster, i.e., the average of all parameters belonging to it. Therefore, the quantization function is
\begin{equation}
Q(v_i)=\frac{\sum_j{\mathbbm{1}\{\hat{Q}(v_j)=\hat{Q}(v_i)\}v_j}}{\sum_j{\mathbbm{1}\{\hat{Q}(v_j)=\hat{Q}(v_i)\}}},
\end{equation}
where $\mathbbm{1}\{statement\}$ equals to $1$ when the statement is true, otherwise $0$.

\subsection{K-Means Quantization}\label{k-means}
\indent Suppose that we need to quantize a vector $v$ to $k$ bits ($k$-bit quantization). For k-means quantization, we leverage the k-means clustering with k-means++ initialization to partition the vector $v$ into $2^k$ clusters.\\
\indent We first utilize k-means++ initialization method to initialize the $2^k$ centroids ($\mu_1,$ $\mu_2,\ ...\ ,\ \mu_{2^k}$) for each cluster ($c_1,\ c_2,\ ...\ ,\ c_{2^k}$). Then, each parameter $v_i$ is classified into its nearest cluster. After all the parameters in $v$ are classified, the centroids are updated as the average of all the parameters that belong to them respectively. Then, repeat re-classifying parameters and updating centroids until convergence is met or the maximum iteration is reached. 
Moreover, the procedure of k-means++ initialization method is as follows: first, choose a random parameter from the vector $v$ as the first centroid; then assign the possibilities to become the next centroids of other parameters according to their smallest distance from all the existing centroids and choose the next centroid based on these possibilities; finally, repeat possibility assignment and centroid choosing until all the $2^k$ centroids are generated. \\

\begin{algorithm}
\caption{k-means clustering}
\begin{algorithmic}[1]
\Require $\#$bits $k$, vector $v$
\Ensure the $2^k$ centroids and the corresponding label vector
\State Initial the $2^k$ centroids
\Repeat
\State Calculate the distance between each parameter $v_i$ and each centroid $\mu_i$ as $d_{i,j}$
\State Classify each parameter $v_i$: $v_i\in c_k$ where $k=\arg\min_j{d_{i,j}}$
\State Update each centroid $\mu_j$: $\mu_j=\frac{\sum_i{\mathbbm{1}\{v_i\in c_j\}v_i}}{\sum_i{\mathbbm{1}\{v_i\in c_j\}}}$
\Until{convergence is met or the maximum iteration is reached}
\end{algorithmic}
\end{algorithm}

\indent To reduce the efficiency drop caused by the upgrading of the quantization scheme, we set the maximum iteration of k-means clustering to only 3. After k-means clustering is finished, We utilize the resulting label vector as the cluster index vector and the resulting centroids as the corresponding centroid value vector. Each parameter $v_i$ will be replaced by the centroid of the cluster which it belongs to. 

\section{Experiments}
\indent In this section, we first introduce the dataset we used in our experiments (Section \ref{dataset}), then explain the experimental details of our experiments on BERT and ALBERT (Section \ref{exp_setup}), finally show the results and the corresponding discussion (Section \ref{results&dicuss}).

\subsection{Dataset}\label{dataset}
\indent We test the performance of our quantized models on the General Language Understanding Evaluation (GLUE) benchmark \cite{glue}. which contains NLU tasks including question answering, sentiment analysis, and textual entailment. Specifically, we utilize 8 tasks (QNLI, CoLA, RTE, SST-2, MRPC, STS-B, MNLI, and QQP) to test the performance of different quantization schemes. 
The evaluation metrics of each task are as follows: Matthews correlation coefficient (mcc) for CoLA; accuracy (acc) for QNLI, RTE, SST-2, and MNLI; accuracy (acc) and F1 score for MRPC and QQP; Pearson and Spearman correlation coefficients (corr) for STS-B. We follow the default split of the dataset. The datasets are available for download here: \href{https://gluebenchmark.com/tasks}{https://gluebenchmark.com/tasks}. 

\begin{table}
\centering
\caption{The results of fixed-precision linear quantization for BERT on GLUE benchmark.}\label{table:bert_linear}
\begin{tabular}{c|c|c|c|c|c|c|c|c|c}
 \toprule[2pt]
 \#bits & QNLI & CoLA & RTE & SST-2 & MRPC & STS-B & MNLI-m/mm & QQP & average\\
 \hline \hline
32 bits & 91.7 & 59.2 & 72.2 & 93.1 & 86.3/90.4 & 89.7 & 85.0/84.8 & 91.6/88.8 & 83.7\\
\hline
5 bits & 88.5 & 48.4 & 69.3 & 89.6 & 83.8/88.7 & 88.7 & 79.8/80.4 & 88.9/85.3 & 79.7\\ 
4 bits & 81.8 & 19.9 & 57.0 & 81.4 & 75.7/84.5 & 84.9 & 71.4/71.9 & 80.8/75.9 & 69.4\\
3 bits & 61.3 & 11.9 & 56.3 & 78.9 & 70.8/81.9 & 68.6 & 59.6/61.6 & 76.5/71.1 & 60.6\\
2 bits & 60.7 & 6.6  & 55.2 & 77.9 & 69.6/81.4 & 47.4 & 49.6/50.8 & 74.2/63.2 & 54.7\\
1 bit  & 59.5 & 0    & 54.9 & 77.5 & 69.9/81.4 & 37.8 & 47.3/48.8 & 74.3/63.3 & 52.2\\

 \bottomrule[1.5pt]
\end{tabular}

\end{table}
\begin{table}
\centering
\caption{The results of fixed-precision k-means quantization for BERT on GLUE benchmark.}\label{table:bert_kmeans}
\begin{tabular}{c|c|c|c|c|c|c|c|c|c}
 \toprule[2pt]
 \#bits & QNLI & CoLA & RTE & SST-2 & MRPC & STS-B & MNLI-m/mm & QQP & average\\
 \hline \hline
32 bits & 91.7 & 59.2 & 72.2 & 93.1 & 86.3/90.4 & 89.7 & 85.0/84.8 & 91.6/88.8 & 83.7\\
\hline
5 bits & 91.5 & 60.2 & 70.8 & 94.0 & 87.3/91.0 & 89.6 & 84.7/84.9 & 91.7/88.8 & 83.9\\ 
4 bits & 91.7 & 57.4 & 70.8 & 93.6 & 87.0/91.0 & 89.6 & 84.8/84.8 & 91.6/88.7 & 83.5\\
3 bits & 91.3 & 56.9 & 70.0 & 93.1 & 86.0/90.2 & 89.4 & 84.4/84.1 & 91.2/88.1 & 82.9\\
2 bits & 89.5 & 50.2 & 66.1 & 91.3 & 84.6/89.2 & 88.3 & 81.6/81.9 & 90.3/87.0 & 80.4\\
1 bit  & 62.2 & 13.7 & 54.5 & 83.0 & 70.8/81.7 & 52.2 & 62.0/62.6 & 77.1/65.9 & 59.8\\
 \bottomrule[1.5pt]
\end{tabular}

\end{table}

\begin{table}
\centering
\caption{The results of fixed-precision linear quantization for ALBERT on GLUE benchmark.}\label{table:albert_linear}
\begin{tabular}{c|c|c|c|c|c|c|c|c|c}
 \toprule[2pt]
 \#bits & QNLI & CoLA & RTE & SST-2 & MRPC & STS-B & MNLI-m/mm & QQP & average\\
 \hline \hline
32 bits & 91.5 & 58.9 & 81.6 & 92.8 & 90.2/93.1 & 90.9 & 84.9/85.1 & 90.8/87.7 & 85.2\\
\hline
5 bits & 60.1 & 0 & 53.1 & 74.8 & 68.4/81.2 & 39.9 & 43.6/45.6 & 72.6/65.8 & 50.9\\ 
4 bits & 52.3 & 0 & 52.7 & 50.9 & 68.4/81.2 & 6.8  & 35.5/35.2 & 67.9/56.5 & 41.1\\
3 bits & 51.4 & 0 & 54.2 & 54.9 & 68.4/81.2 & 16.7 & 35.5/35.4 & 68.2/56.7 & 42.7\\
2 bits & 54.0 & 0 & 52.7 & 50.9 & 68.4/81.2 & 18.8 & 35.4/35.3 & 67.5/53.2 & 42.6\\
1 bit  & 54.3 & 0 & 55.6 & 50.9 & 68.4/81.2 & 9.7  & 35.5/35.3 & 67.3/52.5 & 41.9\\

 \bottomrule[1.5pt]
\end{tabular}

\end{table}
\begin{table}
\centering
\caption{The results of fixed-precision k-means quantization for ALBERT on GLUE benchmark.}\label{table:albert_kmeans}
\begin{tabular}{c|c|c|c|c|c|c|c|c|c}
 \toprule[2pt]
 \#bits & QNLI & CoLA & RTE & SST-2 & MRPC & STS-B & MNLI-m/mm & QQP & average\\
 \hline \hline
32 bits & 91.5 & 58.9 & 81.6 & 92.8 & 90.2/93.1 & 90.9 & 84.9/85.1 & 90.8/87.7 & 85.2\\
\hline
5 bits & 91.0 & 55.9 & 78.3 & 92.7 & 90.7/93.4 & 90.8 & 84.2/85.1 & 90.3/87.1 & 84.3\\ 
4 bits & 90.1 & 48.9 & 75.5 & 87.0 & 84.8/89.3 & 75.8 & 82.1/83.1 & 89.2/85.5 & 79.6\\
3 bits & 63.5 & 4.6  & 53.8 & 76.5 & 68.1/80.8 & 77.7 & 63.7/65.8 & 82.9/77.9 & 61.8\\
2 bits & 61.4 & 0    & 59.9 & 71.6 & 70.8/82.2 & 20.4 & 45.0/45.6 & 72.7/61.5 & 49.7\\
1 bit  & 50.6 & 0    & 56.0 & 52.2 & 68.4/81.2 & 6.3  & 35.4/35.2 & 69.8/58.8 & 41.5\\
 \bottomrule[1.5pt]
\end{tabular}

\end{table}

\subsection{Experimental Setup}\label{exp_setup}
\indent Before quantization, the bert-base-uncased version of BERT models is fine-tuned on the 8 tasks by the Adam optimizer \cite{adam} and the linear schedule with a learning rate of 5e-5. As for ALBERT models, We first fine-tune the albert-base-v2 model on QNLI, CoLA, SST-2, MNLI, and QQP, and then further fine-tuned on RTE, MRPC, and STS-B basing on the MNLI checkpoint (following the same process as \cite{albert}). We use Adam optimizer and linear schedule to fine-tune ALBERT, and the learning rate for each tasks is searched in $\{$1e-5, 2e-5, 3e-5, 4e-5, 5e-5$\}$.

\indent After quantization, we further fine-tune the quantized models on the corresponding tasks.
In particular, the learning rates of the layers which are quantized are multiplied 10 times (i.e., 5e-4 for all the quantized BERT models) while those of other layers remained the same.


\subsection{Experimental Results and Discussion}\label{results&dicuss}
\begin{figure}
    \centering
    \includegraphics[width=330pt]{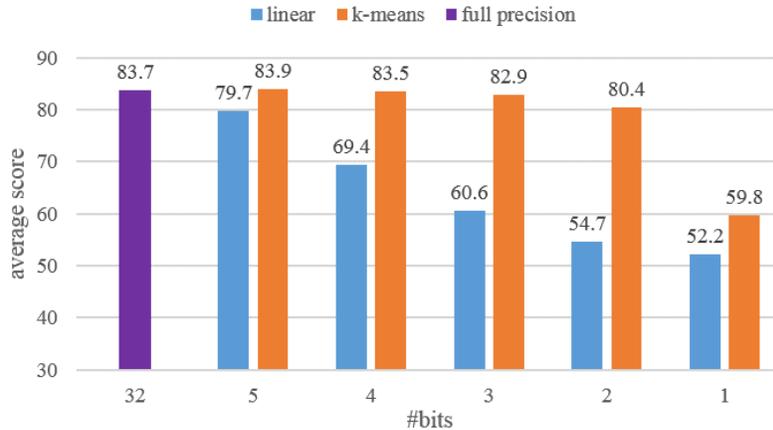}
    \caption{The comparison of average scores of the 8 GLUE tasks for linear and k-means quantization on BERT models.}
    \label{fig:bert}
\end{figure}

\begin{figure}
    \centering
    \includegraphics[width=330pt]{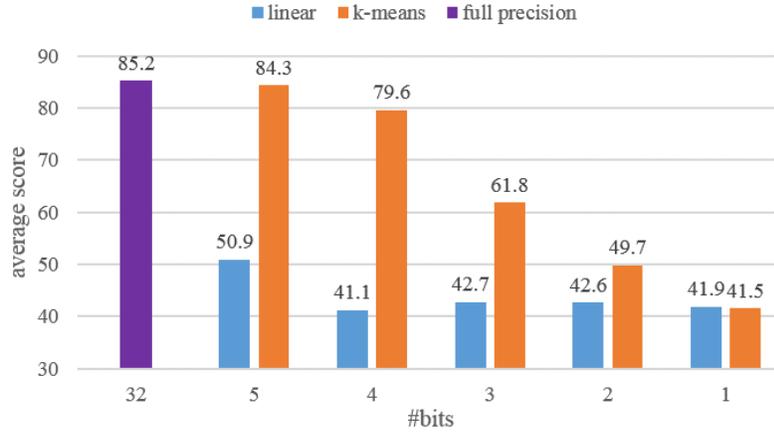}
    \caption{The comparison of average scores of the 8 GLUE tasks for linear and k-means quantization on ALBERT models. }
    \label{fig:albert}
\end{figure}

\indent We mainly focus on 1-5 bits fixed-precision quantization. The results of linear and k-means quantization for BERT are shown in Table \ref{table:bert_linear} and Table \ref{table:bert_kmeans} respectively, and further comparison between the average scores of the two sets of experiments is shown in Figure \ref{fig:bert}. Similarly, The results and comparison of ALBERT are shown in Table \ref{table:albert_linear}, Table \ref{table:albert_kmeans}, and Figure \ref{fig:albert} respectively.

\subsubsection{BERT}
\indent \textbf{The improvements brought by quantization scheme upgrading.} As shown in Table \ref{table:bert_linear}, Table \ref{table:bert_kmeans} and Figure \ref{fig:bert}, although the models perform worse with lower bits no matter which quantization scheme is utilized, the models quantized with k-means quantization perform significantly better than those using linear quantization in each bit setting respectively, across all 8 tasks and their average. On average of 8 tasks, only by upgrading quantization scheme from linear to k-means, we achieve a performance degradation drop from (38.8$\%$, 34.7$\%$, 27.6$\%$, 17.1$\%$, 4.8$\%$) to (28.6$\%$, 3.94$\%$, 0.9$\%$, 0.3$\%$, -0.2$\%$) for 1-5 bits quantization respectively, as compared to the full precision model. The result shows that great performance improvements could be achieved by only upgrading the quantization scheme, which indicates that the improvement space of the quantization scheme is much underestimated. To further illustrate it, we repeated several experiments using the group-wise linear quantization scheme developed by \cite{q-bert} which is an improvement based on linear quantization and achieves much higher performance than simple linear quantization. The results are shown in Table \ref{table:group}. Compared to the performance of group-wise linear quantization, simple k-means quantization achieve even higher performance or comparable performance while saving a huge amount of time.$\footnote{In group-wise quantization, each matrix is partitioned to different groups and each group is quantized separately. For the forward pass, the model needs to reconstruct each quantized group respectively for each layer instead of reconstructing the entire weight matrix of each quantized layer directly. That explains why group-wise quantization is quite time-consuming. Specifically, in our group-wise quantization experiments, we partition each matrix to 128 groups.}$


\begin{table}
\centering
\caption{The comparison between k-means quantization and group-wise linear quantization on BERT. The rightmost column are the average accelerations of k-means quantization compared to group-wise linear quantization on RTE and MRPC. The experiments are carried out using four NVIDIA 2080 Ti. }\label{table:group}
\begin{tabular}{c|c|c|c}
 \toprule[2pt]
 Model & RTE & MRPC & acceleration \\
 \hline \hline
3 bits k-means & 70.0 & 86.0/90.2 & \multirow{2}{*}{22$\times$} \\
3 bits group-wise & 72.6 & 84.8/89.6 &\\
\hline
2 bits k-means & 66.1 & 84.6/89.2 & \multirow{2}{*}{16$\times$} \\
2 bits group-wise & 58.5 & 72.3/81.1 &\\
\hline
1 bit k-means & 54.5 & 70.8/81.7 & \multirow{2}{*}{10$\times$} \\
1 bit group-wise & 53.1 & 70.6/81.4 & \\
 \bottomrule[1.5pt]
\end{tabular}

\end{table}

\indent \textbf{The potential of k-means quantization.} As shown in Table \ref{table:bert_kmeans}, the model can be compressed well simply using k-means quantization with fixed-precision strategies, and the quantized models still perform well even in some particularly low bit settings. For instance, on the task RTE, the model quantized to 3 bits with k-means quantization only results in a 2.16 $\%$ performance degradation. For most tasks including QNLI, SST-2, MRPC, STS-B, MNLI, and QQP, the performance of the quantized models only show a significant drop in 1-bit setting. It is worth noting that these results were achieved by simple k-means quantization with a maximum iteration of only 3 and without any tricks, which indicates the great developing potential of k-means quantization. 


\subsubsection{ALBERT}
\indent Generally speaking, the two main arguments drew from BERT experiments still hold as shown in Table \ref{table:albert_linear}, Table \ref{table:albert_kmeans}, and Figure \ref{fig:albert}. We could also see great improvements brought by quantization scheme upgrading
and great potential of k-means quantization.
However, there are some abnormal results which are worth 
discussing.

\begin{table}
\centering
\caption{The performance of 1-bit quantization with different number of k-means iteration on ALBERT.}\label{table:iteration}
\begin{tabular}{c|c|c|c}
 \toprule[2pt]
 Iteration & QNLI & MRPC & STS-B \\
 \hline \hline
3  & 50.56 & 68.38/81.22 & 6.29  \\
5  & 50.63 & 68.38/81.22 & 6.93  \\
10 & \textbf{60.63} & 68.87/81.30 & \textbf{13.76} \\
20 & 60.19 & \textbf{69.85/81.83} & 11.10 \\
 \bottomrule[1.5pt]
\end{tabular}

\end{table}

\indent \textbf{The influence of the number of k-means iterations.} The first set of abnormal results is from 1-bit quantization of QNLI, MRPC, and STS-B. While k-means normally outperformed linear quantization, these results violate this regulation. We believe that is because the distribution of parameters is so complicated that 3 iterations of k-means could not work well. To validate this theory and further explore the influence of iterations, we repeated the experiments with these abnormal results while extending the number of iteration to 5, 10, and 20. The corresponding results are shown in Table \ref{table:iteration}. With more iterations, the accuracy of k-means quantization increases and outperforms linear quantization. However, the over-fitting problem might be troublesome as the performances decrease for QNLI and STS-B when the number of iteration increases from 10 to 20. Therefore, in k-means quantization, 
the number of k-means iterations is also an important hyper-parameter that needs to be searched carefully.

\indent \textbf{The special number of CoLA and MRPC.} Another set of abnormal results is from the linear quantization of CoLA and MRPC, which are binary classification tasks. We find the quantized models output ``1'' all the time after being fine-tuned. The two values $0$ and $68.4$ are only determined by the data distribution on the dev sets. In other words, after the model is quantized to 1-5 bits with linear quantization, it almost loses its functionality and becomes difficult to train on the two tasks. Moreover, we further do experiments in high bit settings on the two tasks and find that the results of the quantized models are no longer the two values starting from 6 bits. 



\begin{figure}
    \centering
    \includegraphics[width=330pt]{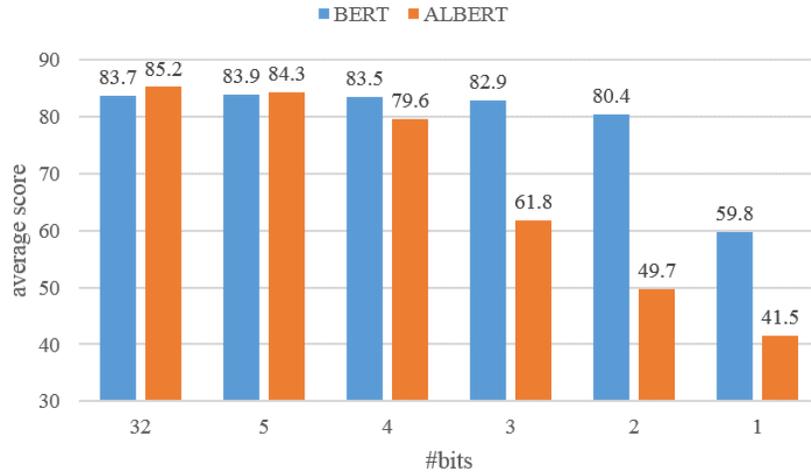}
    \caption{The comparison of average scores of the 8 GLUE tasks for BERT and ALBERT models with k-means quantization.}
    \label{fig:comparison}
\end{figure}

\begin{figure}
    \centering
    \includegraphics[width=330pt]{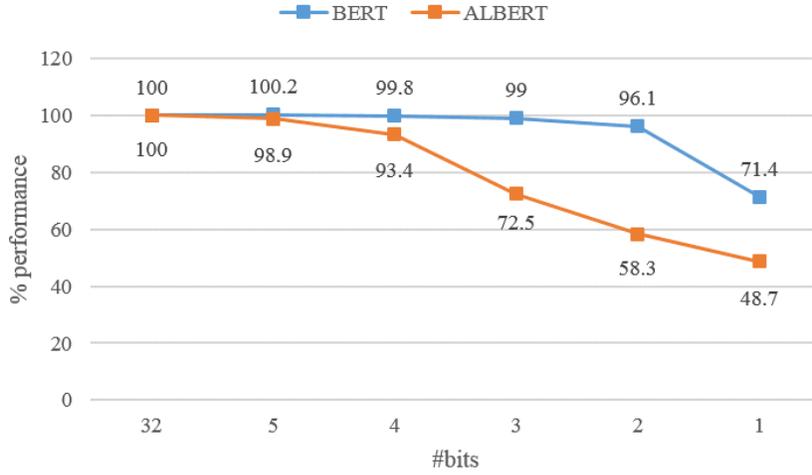}
    \caption{The comparison of performance for BERT and ALBERT models with k-means quantization. Each value refers to the percentage of the average score of the quantized model compared to the score of the full precision model.}
    \label{fig:comparison_1}
\end{figure}

\indent \textbf{The comparison between BERT and ALBERT.} Moreover, we compare the performances between k-means quantization for BERT and ALBERT, and the results are shown in Figure \ref{fig:comparison} and Figure \ref{fig:comparison_1}. 
Compared with BERT which remains 96.1\% of its origin performance after k-means 2-bit quantization, ALBERT is much less robust in terms of quantization (in our work, robustness towards quantization means the ability to quantize to low bit-width while maintaining high performance). The performance of ALBERT falls to 93.4\% and 72.5\% after k-means 4-bit and 3-bit quantization respectively. Consider that the major improvement of ALBERT based on BERT is parameter sharing and quantization can also be considered as intra-layer parameter sharing, we speculate that parameter sharing and quantification have similar effects, which means that the redundant information removed by parameter sharing and quantization partially overlaps. Moreover, after parameter sharing, ALBERT has removed a great amount of redundant information compared to BERT (the total number of parameters fall from 108M to 12M). Therefore, further applying quantization upon ALBERT will easily damage the useful information and the robustness of ALBERT towards quantization is rather low. 
However, from another point of view, the parameter sharing has already significantly reduced the parameter number and thus can also be considered as a model compression method. Moreover, consider that the performances of full-precision ALBERT are better than those of 4-bit and 3-bit BERT models which occupy a similar amount of memory in GPU, the parameter sharing can even achieve better compress performance than simple quantization. However, as a compression method, parameter sharing has a non-negligible drawback: it can only reduce the memory consumption while most other compression methods can reduce both the memory consumption and the calculation consumption (i.e. the inference time).


\section{Conclusion}
\indent In this paper, we compare k-means and linear quantization on BERT and ALBERT models and get three main results. First, we find the models quantized with k-means significantly outperform those using linear quantization. Great performance improvements could be achieved by simply upgrading the quantization scheme. 
Second, the model can be compressed to relatively low bit-width only using k-means quantization even with simple fix-precision strategy and without any tricks. That indicates the great developing potential of k-means quantization.
Third, the number of k-means iterations plays an important role in the performance of quantized models and should be determined carefully.
Besides, through comparison between the results of k-means quantization for BERT and ALBERT, we discover that ALBERT is much less robust towards quantization than BERT. That indicates that parameter sharing and quantization have some effects in common. Therefore, further applying quantization upon models with extensive parameter sharing will easily damage the useful information and thus lead to a significant performance drop.

\subsubsection{Acknowledgement} We thank the anonymous reviewers for their thoughtful comments. This work has been supported by the National Key Research and Development Program of China (Grant No. 2017YFB1002102) and Shanghai Jiao Tong University Scientific and Technological Innovation Funds (YG2020YQ01).

%
%
%
\bibliographystyle{splncs04}
\bibliography{references}

\end{document}